## Appendix A. Supplementary Material

### A.1. Properties of Data

Table 5 shows the number of reviews per domain for the balanced dataset. Note that the data was downloaded up until July 2011, and there may be variations in the names of the categories between then and now.

### A.2. SSL

Figure 11 shows the attempt at combining the random and highest-margin selection methods, as discussed in section 6.

### A.3. Domain Adaptation

In Figures 5 to 8 we present the one-to-one adaptation in settings 1 and 3 (mixed training set), discussed in section 7. Figure 12 shows the percentage of 10 domains used in the 1-to-many SSL. Only the top 5 and least 5 frequent domains are shown. Note that since the industrial domain is the source domain, and it does not appear in the unlabeled set, it begins (and remains) at a 100%.

### A.4. WSL

The results for the experiments discussed in section 8 are shown in figure 10. Example book reviews are shown in figure 9 (with grammar and spelling mistakes intact).

17

| Domain | # of Reviews | Test Set Size |
|---|---:|---:|
| 1. books | 1,713,900 | 100,000 |
| 2. movies (Movies & TV) | 555,208 | 10,000 |
| 3. elect (Electronics) | 426,258 | 10,000 |
| 4. music | 293,704 | 10,000 |
| 5. kindle (Kindle Store) | 226,752 | 10,000 |
| 6. videos (Amazon Instant Videos) | 138,228 | 10,000 |
| 7. kitchen (Kitchen & Dining) | 133,740 | 10,000 |
| 8. health (Health & Personal Care) | 107,336 | 10,000 |
| 9. mp3 (MP3 Downloads) | 101,468 | 10,000 |
| 10. video_games | 88,360 | 1,000 |
| 11. home (Home Improvement) | 77,656 | 1,000 |
| 12. sports (Sports & Outdoors) | 77,030 | 1,000 |
| 13. toys (Toys & Games) | 76,388 | 1,000 |
| 14. garden_pets (Home, Garden & Pets) | 72,542 | 1,000 |
| 15. clothing (Clothing & Accessories) | 65,860 | 1,000 |
| 16. beauty | 58,156 | 1,000 |
| 17. baby | 56,994 | 1,000 |
| 18. camera (Camera & Photo) | 46,228 | 1,000 |
| 19. food (Grocery & Gourmet Food) | 45,270 | 1,000 |
| 20. software | 40,954 | 1,000 |
| 21. shoes | 36,986 | 1,000 |
| 22. cell_phones (Cell Phones & Accessories) | 36,100 | 1,000 |
| 23. patio (Patio, Lawn & Garden) | 33,194 | 1,000 |
| 24. office (Office Products) | 17,958 | 1,000 |
| 25. auto (Automotive) | 17,756 | 1,000 |
| 26. computer (Computer & Accessories) | 14,544 | 1,000 |
| 27. watches | 12,960 | 1,000 |
| 28. musical_inst (Musical Instruments) | 9,254 | 1,000 |
| 29. android (Appstore for Android) | 7,452 | 1,000 |
| 30. jewelry | 7,026 | 1,000 |
| 31. magazine (Magazine Subscriptions) | 4,902 | 100 |
| 32. arts (Arts, Crafts & Sewing) | 4,084 | 100 |
| 33. industrial (Industrial & Scientific) | 1,142 | 100 |
| **Total** | **4,605,454** | |

Table 5: Properties of the balanced data.

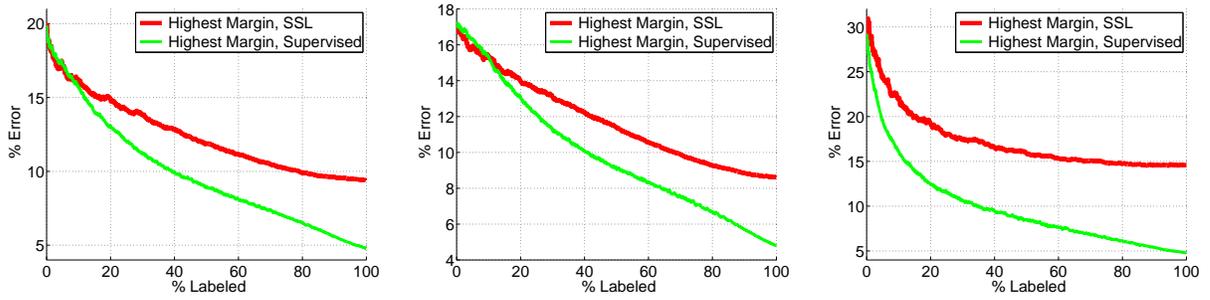

Figure 5: DA. left: Movies to Books. middle: Movies&Books to Books. right: Elect. to Books

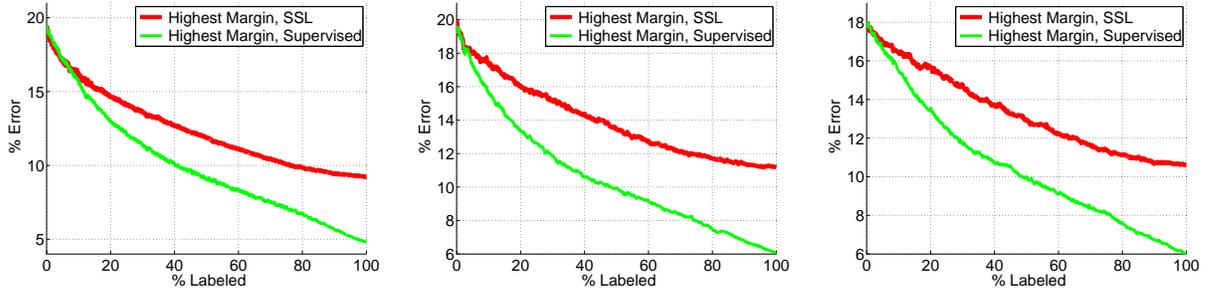

Figure 6: DA. left: Elect.&Books to Books. middle: Books to Movies. right: Books&Movies to Movies

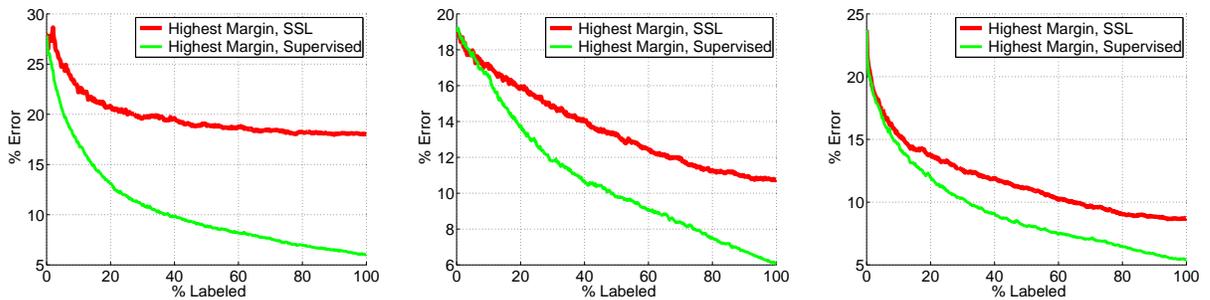

Figure 7: DA. left: Elect. to Movies. middle: Elect.&Movies to Movies. right: Movies to Elect.

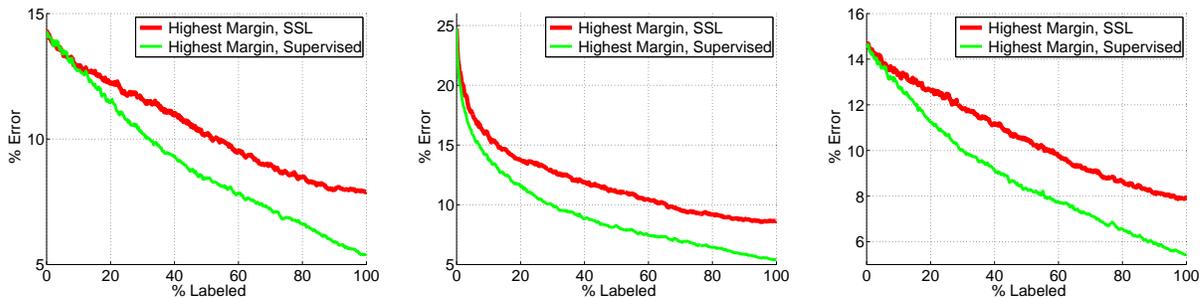

Figure 8: DA. left: Movies&Elect to Elect. middle: Elect.&Movies to Movies. right: Movies to Elect.

| excellent story | great fiction | poor | It's horrible!!!! |
|---|---|---|---|
| i usually get my books from the library but this one was so good i wanted to get my own copy!!! excellent story! | There's nothing really scientific on this book. It's only a great fiction work really. Almost comic most of the time. Have that in mind when you read it! | sorry Barbara i usally love your books but this one is really poor. the ending is a mess there is no character developement no scene painting ... intreaging. | It is so horrible to believe that this truely happens to children and their parents. I wish I had read the book yrs. ago but thank the lord my children were okay. |

Figure 9: Left to right: Example of true positive, false positive, true negative and false negative reviews labeled using the WSL rules.

20

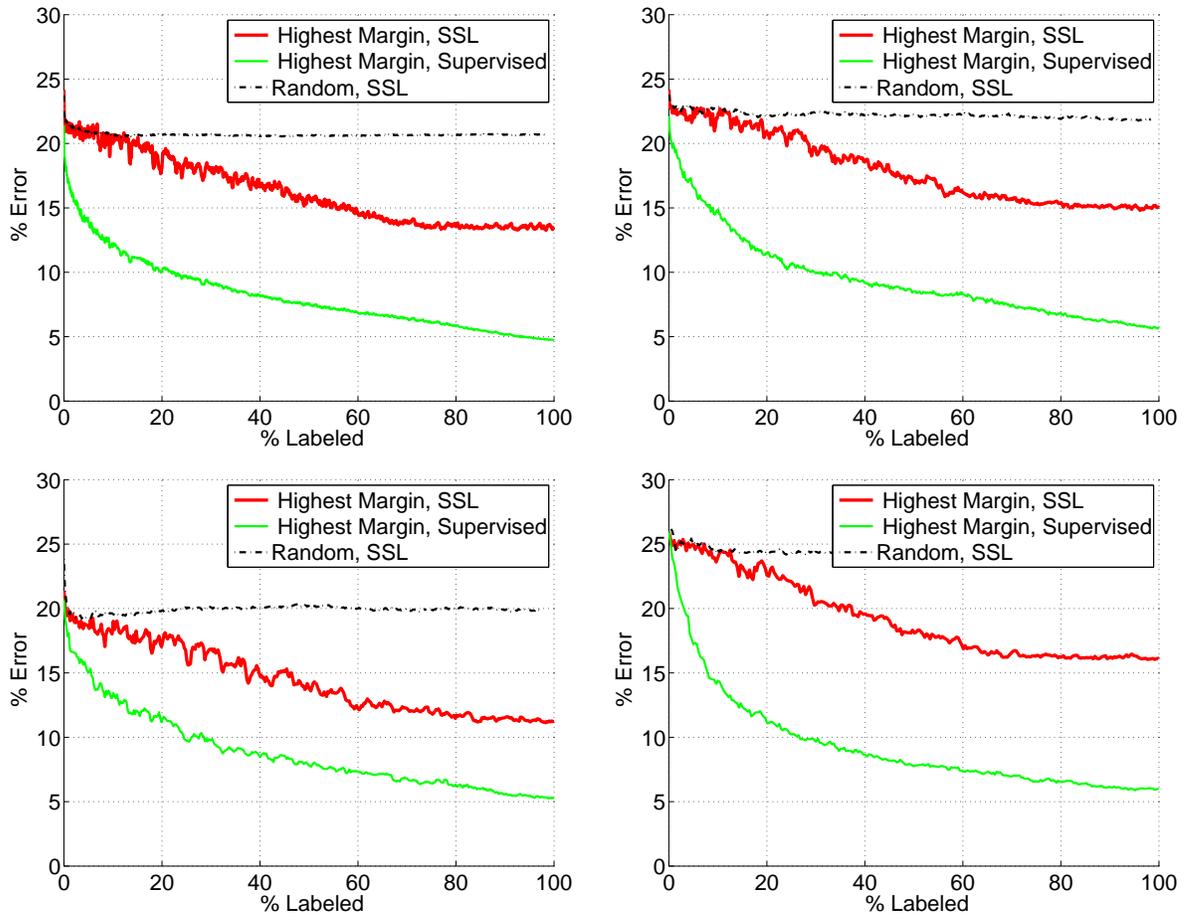

Figure 10: WSL Error rate. top left: Books. top right: Movies. bottom left: Electronics. bottom right: Music.

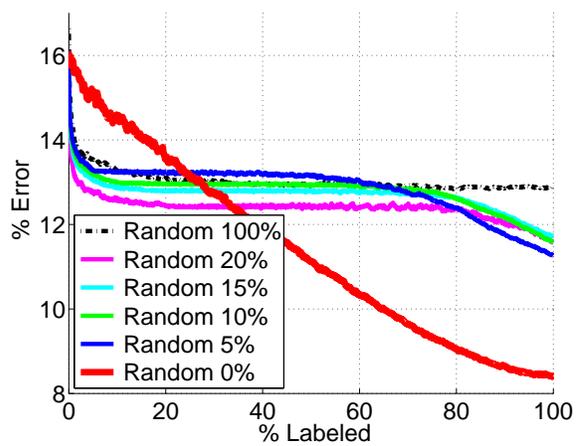

Figure 11: Starting with random sampling before changing to highest-margin.

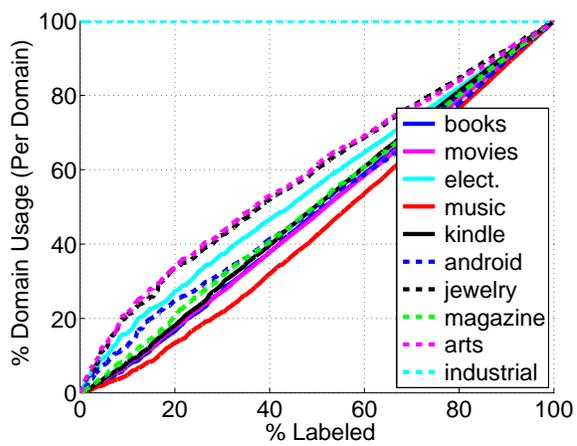

Figure 12: Domain usage of DA from industrial to 32 other domains

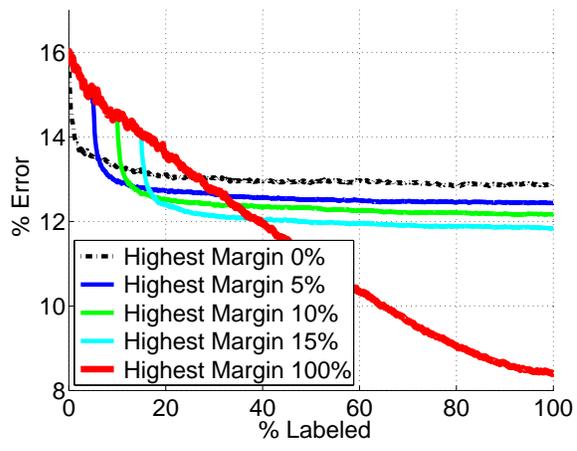

Figure 13: Starting with highest-margin before changing to random sampling.

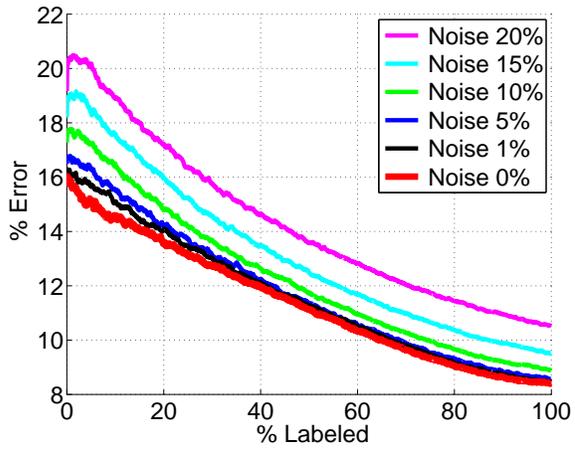

Figure 14: Initial label noise

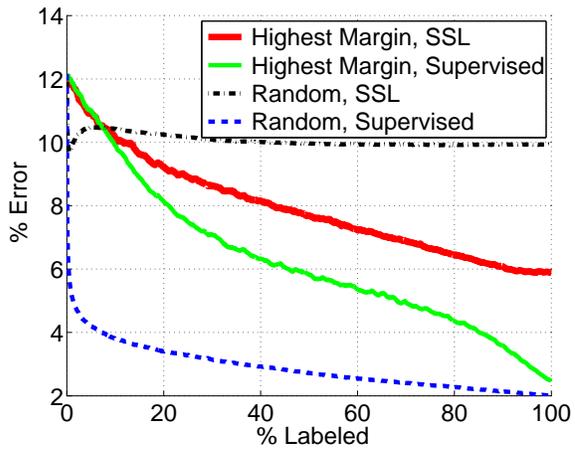

Figure 15: SSL on 6M book reviews (85% positive)

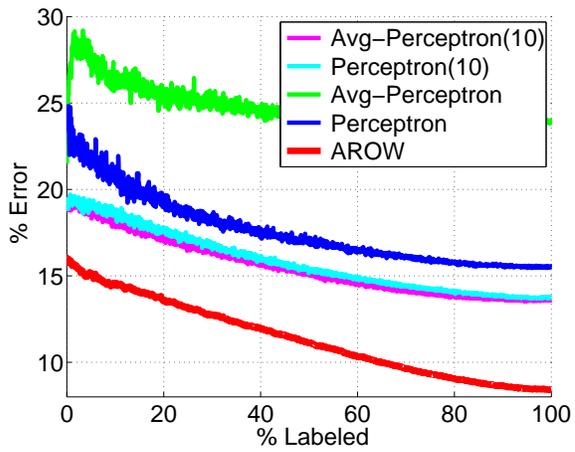

Figure 16: AROW vs. Perceptron

23